\pdfoutput=1
\documentclass[11pt]{article}

\usepackage[]{acl}

\usepackage{times}
\usepackage{latexsym}
\usepackage{enumitem}

\usepackage[T1]{fontenc}
\usepackage[utf8]{inputenc}
\usepackage{amsmath,amsfonts,bm}
\usepackage{microtype}

\usepackage{todonotes}
\usepackage{booktabs}
\usepackage{graphicx}
\usepackage{subcaption} % For subtables
\usepackage{fontawesome} % For weird symbols 
\usepackage{comment}

\newcommand{\politifact}{PolitiFact}
\newcommand{\vclaim}{\textit{VerClaim}}
\newcommand{\inputclaim}{\textit{InputClaim}}
\newcommand{\clean}{\textit{clean}~}
\newcommand{\cleanhard}{\textit{clean-hard}}
\newcommand{\partof}{\textit{part-of}}
\newcommand{\contextdep}{\textit{context-dep}}
\newcommand{\featurecontext}{\textit{\textbf{FC}}}

\newcommand{\concat}{\ensuremath{+\!\!\!\!+\,}}

\title{The Role of Context in Detecting Previously Fact-Checked Claims}

\author{
    Shaden Shaar$^1$, Firoj Alam$^1$, Giovanni Da San Martino$^2$, Preslav Nakov$^1$\\
  $^1$Qatar Computing Research Institute, HBKU, Doha, Qatar \\
  $^2$University of Padova, Italy \\
  \texttt{\{sshaar, falam, pnakov\}@hbku.edu.qa}\\\texttt{dasan@math.unipd.it} 
  \\}

\begin{document}
\maketitle
\begin{abstract}
Recent years have seen the proliferation of disinformation and fake news online. Traditional approaches to mitigate these issues is to use manual or automatic fact-checking. Recently, another approach has emerged: checking whether the input claim has previously been fact-checked, which can be done automatically, and thus fast, while also offering credibility and explainability, thanks to the human fact-checking and explanations in the associated fact-checking article. Here, we focus on claims made in a political debate and we study the impact of modeling the context of the claim: both on the source side, i.e., in the debate, as well as on the target side, i.e., in the fact-checking explanation document. We do this by modeling the local context, the global context, as well as by means of co-reference resolution, and multi-hop reasoning over the sentences of the document describing the fact-checked claim. The experimental results show that each of these represents a valuable information source, but that modeling the source-side context is most important, and can yield 10+ points of absolute improvement over a state-of-the-art model.
\end{abstract}

\section{Introduction}
\label{sec:introduction}

The fight against dis/mis-information has become an urgent social and political matter. Online media have been widely used not only for social good, but also to mislead entire communities. 
Many fact-checking organizations, such as FactCheck.org,\footnote{\url{http://www.factcheck.org/}}
Snopes,\footnote{\url{http://www.snopes.com/fact-check/}}
PolitiFact,\footnote{\url{http://www.politifact.com/}}
and FullFact,\footnote{\url{http://fullfact.org/}}
as well as some broader international initiatives such as the \textit{Credibility Coalition}\footnote{\url{https://credibilitycoalition.org/}} 
and \textit{Eufactcheck},\footnote{\url{https://eufactcheck.eu/}} 
have emerged to address the problem \cite{stencel2019number}.

There have also been efforts to develop automatic systems to detect such content \cite{vo2018rise,Shu:2017:FND:3137597.3137600,thorne-vlachos:2018:C18-1,Li:2016:STD:2897350.2897352,Lazer1094,Vosoughi1146,CIKM2020:FANG}, including the development of datasets \cite{augenstein-etal-2019-multifc}, systems \cite{WhatTheWikiFact}, and evaluation campaigns \cite{clef-checkthat-lncs:2020,nakov2021clef,clef-checkthat:2021:LNCS,shaar2021findings,CheckThat:ECIR2022}. 

An important issue with automatic systems is that journalists and fact-checkers often question their credibility for reasons such as (perceived) insufficient accuracy given the state of present technology, but also due to the lack of explanation about how the system has made its decision. On the other hand, manual fact-checking is time-consuming and does not scale. Yet, time is precious: it has been reported in the literature that \textit{fake news} travels faster than real news \cite{vosoughi2018spread}, and that 50\% of the spread of some very viral false claims has happened within the first ten minutes after they got published \cite{zaman2014bayesian}. Such findings show the importance of real-time fake news detection, which can enable a timely intervention.

As both manual and automatic systems have their limitations, there have been proposals for human-in-the-loop settings, aiming to bring the best of both worlds. In order to enable such an approach, one question that arises is how to facilitate fact-checkers and journalists with automated systems \cite{Survey:2021:AI:Fact-Checkers}. An immediate problem is to know whether a given input claim has been previously fact-checked by a reputable fact-checking organization. This would give the journalist a credible reference and could save her significant amount of time, as manually fact-checking a single non-trivial claim may take from 1-2 days to 1-2 weeks. While earlier studies have suggested that such a mechanism should be part of an end-to-end automated system, there has been limited work in this direction \cite{shaar-etal-2020-known,vo-lee-2020-where-facts}.

\begin{figure}[tbh]
\centering
\includegraphics[width=0.82\linewidth]{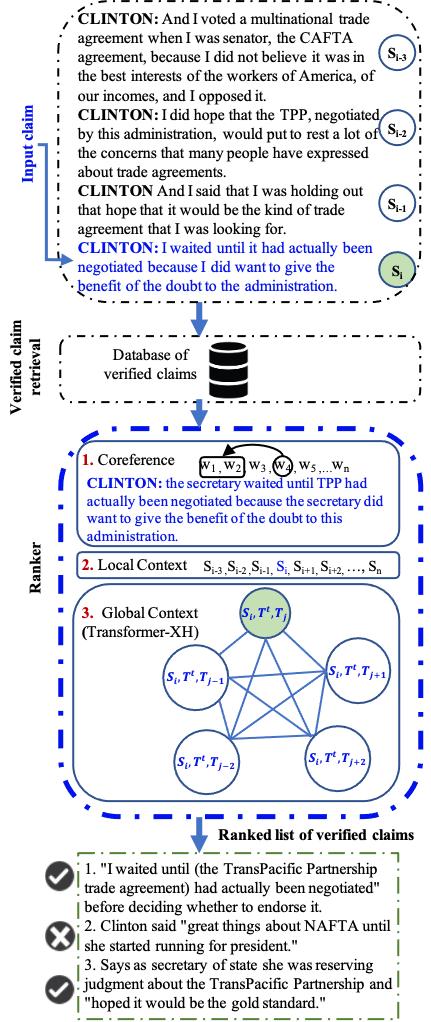}
\caption{A pipeline of retrieving and ranking previously fact-checked claims. $S_{i}$ is the claim (source), $T^{t}$ is the title of the target, $T_{j}$ is a sentence from the target.}
\label{fig:previously_fack_check_pipeline}
\end{figure}

At the time of COVID-19, there are a number of false claims and conspiracy theories spreading online, e.g.,~about Bill Gates and his chips in the COVID-19 vaccine, about garlic water as a cure, etc. Many such claims have already been debunked, but this does not stop them, as they keep being repeated, potentially in a slightly different form but with the same meaning. Thus, it is important to recognize such variations quickly, and possibly to post a reply in social media with a link to a fact-checking article. Similarly, in a scenario where a politician is being interviewed or is taking part in a debate, a quick check against a collection of previously fact-checked claims would make it possible to put him/her on the spot in real time.

However, the problem in a real-time scenario is that, unlike written text, interviews, debates, and speeches are more spontaneous, 
and the claims that are being made are often not clearly formulated in a single sentence. This is illustrated in  Figure~\ref{fig:previously_fack_check_pipeline}, where we can see a fragment from a Democratic debate as part of the 2016 US Presidential election, where Hillary Clinton said: ``\emph{I waited until it had actually been negotiated because I did want to give the benefit of the doubt to the administration.}'' Understanding this claim requires pronominal co-reference resolution (e.g.,~what does \emph{it} refer to, is it \emph{CAFTA} or is it \emph{TPP}, as both are mentioned in the previous sentences?), more general co-reference (e.g.,~that the administration being discussed is the \emph{Obama} administration), as well as general understanding of the conversation so far, and possibly general world knowledge about US politics at the time of the debate (e.g.,~that Hillary Clinton was Secretary of State when TPP was being discussed). 

Moreover, previous research has shown that it is beneficial to match the input claim not only against the canonical verified claim that fact-checkers worked with, but against the entire article that they wrote explaining why the claim was ultimately judged to be true/false \cite{shaar-etal-2020-known,vo-lee-2020-where-facts}. This is because, in the fact-checking article, the claim is likely to be paraphrased in different ways, and there could also be background information and related terms, which can facilitate claim matching, and thus improve recall. 
This means that we need to make use of the global contextual information contained within the full text of the fact-checking article or at least the sentences next to the claim, i.e.,~the local context.
Similarly, for the FEVER fact-checking task, which asks to fact-check against Wikipedia, it has been shown that multi-hop reasoning (Transformer-XH) over the sentences of the target article can help \cite{zhao2019transformer}, an observation that was further confirmed in the context of fact-checking political claims \cite{ostrowski2020multi}. Transformer-XH uses a novel attention mechanism that naturally ``hops'' across the connected text sequences in addition to attending over tokens within each sequence. As claims and reasoning about them are manifested across documents, this hop-based attention mechanism constructs global contextualized representation to provide better joint multi-evidence reasoning. In the present work, we rely on Transformer-XH to extract and use global contextual information.

Based on the above considerations, we propose a framework that focuses on modeling co-reference, local context (representation from neighboring sentences; see Section~\ref{ssec:local_context}), and global context (representation from Transformer-XH; see Section~\ref{ssec:global_context}), both on the source and on the target side, while also using multi-hop reasoning over the target side.

Our contributions can be summarized as follows:

\begin{itemize} 
    \item We perform careful manual analysis to understand what makes detecting previously fact-checked claims a hard problem, and we categorize the claims by type. We release these annotations to enable further research.
    \item Unlike previous work, we focus on modeling the context both on the source side and on the target side, both locally and globally, using co-reference resolution and reasoning with Transformer-XH, which yields sizable improvements over state-of-the-art models of over ten MAP points absolute. 
    \item We propose a realistic and challenging, time-sensitive and document-aware, data split compared to previous work, which we also release.\footnote{\url{https://github.com/firojalam/Detecting-Previously-Fact-Checked-Claims.git}}
\end{itemize}

% %%%%%%%paper structure
The rest of the paper is organized as follows. Section~\ref{sec:related_work} provides a brief overview of previous work. Section~\ref{sec:dataset} introduces the dataset development process. Section~\ref{sec:experimental_setup} presents the experiments. 
Section~\ref{sec:results} discusses the evaluation results.
Finally, Section~\ref{sec:conclusion} concludes with lessons learned and points to possible directions for future work.
\section{Related Work}
\label{sec:related_work}

Below, we describe three relevant lines of research: on detecting previously fact-checked claims, on semantic matching and ranking, and on context modeling for factuality.

\subsection{Previously Fact-Checked Claims}

While there is a surge in research on automatic fact-checking, fully automatic systems suffer from credibility issues, e.g.,~in the eyes of journalists, and manual checking is still the norm. Thus, it is important to reduce that manual effort by detecting when a claim has already been fact-checked. 

A recent survey has identified the task of detecting previously fact-checked claims as one of the most important ways in which automation can assist human fact-checkers \cite{Survey:2021:AI:Fact-Checkers}.
The task was recognized as an important element of the typical sequence of fact-checking steps \cite{vlachos2014fact}: (\emph{i})~extracting statements that are to be fact-checked, (\emph{ii})~constructing appropriate questions, (\emph{iii})~obtaining the pieces of evidence from relevant sources, and (\emph{iv})~reaching a verdict using that evidence. \citet{Hassan:2017:CFE:3137765.3137815} also mentioned the task as an important component of their end-to-end fact-checking pipeline, but did not evaluate it as a component on its own right.

Recently, \citet{shaar-etal-2020-known} gave a formulation of the task of detecting previously fact-checked claims, and proposed a learning-to-rank approach combining BM25 retrieval with BERT-based semantic matching. They further developed two specialized datasets: (a)~on political debates and speeches, using fact-checked claims from PolitiFact, and (b)~on tweets, using claims from Snopes.

The CLEF 2020-2022 CheckThat! lab~\citep{CheckThat:ECIR2020,clef-checkthat-ar:2020,clef-checkthat-en:2020,nakov2021clef,clef-checkthat:2021:LNCS,clef-checkthat:2021:task2,CheckThat:ECIR2022,clef-checkthat:2022:LNCS,clef-checkthat:2022:task2} extended these datasets with additional data in English and Arabic, adding more data each year. The best systems \citep{clef-checkthat:2021:task1:nlytics2021,clef-checkthat:2021:task2:DIPS,clef-checkthat:2021:task3:Chernyavskiy2021} used a combination of BM25 retrieval, semantic similarity using sentence embeddings~\cite{reimers-gurevych-2019-sentence}, and reranking. \citet{clef-checkthat-Bouziane:2020} further used external data from fact-checking datasets~\cite{wang:2017:Short,thorne-etal-2018-fever,wadden-etal-2020-fact}. 

\citet{NAACL:2022:Batch:loss} fine-tuning BERT using batch-softmax contrastive loss as an alternative to mean squared error and triplet loss, and demonstrated sizable performance gains for a number of sentence scoring tasks, including detecting previously fact-checked claims.

Another recent work by \citet{sheng-etal-2021-article} highlighted the importance of using lexical, semantic, and pattern-based information and proposed a re-ranker based on memory-enhanced transformers for claim matching.

\citet{vo-lee-2020-where-facts} proposed a multi-modal text+image neural ranking model for detecting previously fact-checked claims about images.

However, none of the above work modeled the context of the input claim, which is our focus here.

\subsection{Semantic Matching and Ranking}
Here, we focus on the textual formulation of the problem, as defined by \citet{shaar-etal-2020-known}: given an input claim, we want to detect potentially matching previously fact-checked claims and to rank them accordingly. A related research area is semantic matching and ranking, as matching some \inputclaim--\vclaim~pairs might require sentence embeddings, natural language inference, and coreference resolution. An example of such a difficult pair is shown in Table~\ref{table:dataset-examples}, line 607.
Recent relevant work has used neural approaches. \citet{nie2019combining} proposed a semantic matching method that combines document retrieval, sentence selection, and claim verification neural models to extract claims and to verify them. \citet{thorne-etal-2018-fever} proposed a simple model, where pieces of evidence are concatenated together and then fed into a Natural Language Inference (NLI) model. 
\citet{yoneda2018ucl} used a four-stage approach that combines document and sentence retrieval with NLI. \citet{hanselowski2018ukp} used a BiLSTM-based enhanced sequential inference model~\cite{chen2016enhanced} to rank candidate facts and to classify a new claim based on the selected facts. Several studies used model combination (i.e.,~document retrieval, sentence retrieval, and NLI to classify the retrieved sentences) with joint learning \cite{yoneda2018ucl,hidey-diab-2018-team,luken2018qed}.

\subsection{Context Modeling for Factuality}

Previous work has shown that modeling the context can help for predicting the check-worthiness of claims in political debates, e.g.,~the interaction between the debaters, and the reaction of the moderator and of the public to what was said \cite{gencheva2017context,atanasova2019automatic,RANLP2019:checkworthiness:multitask}. The CLEF 2018-2022 CheckThat! lab had a shared task on this \citep{clef2018checkthat:task1,clef-checkthat-T1:2019,clef-checkthat-en:2020,clef-checkthat:2021:task1,clef-checkthat:2022:task1}.

The CLEF-2018 CheckThat! lab featured a shared task on fact-checking a claim in the context of a political debate \cite{clef-checkthat-T2:2018,clef2018checkthat}, and SemEval-2019 had a shared task on fact-checking in community question answering forums \cite{mihaylova-etal-2019-semeval}.

\citet{liu-etal-2020-fine} proposed a kernel graph attention network to model evidence as a context for fact verification. 
Similarly, \citet{zhou2019gear} used a fully connected evidence graph with multi-evidence information sources for fact verification. 

\citet{zhong2020reasoning} used different pre-trained Transformer models and a graph-based approach, i.e.,~graph convolutional network and graph attention network, for fact verification. \citet{zhao2019transformer} introduced extra hop attention to incorporate contextual information, while maintaining the Transformer capabilities, thus making it possible to learn a global representation of the different pieces of evidence and to jointly reason over the evidence graph. One of the limitations of their approach was the need for human-labeled evidence in relation to the input claims in existing fact-verification datasets. \citet{ostrowski2020multi} developing a dataset of annotated pieces of evidence associated with input claims and used multihop attention to make a prediction about the factuality of a claim.

Unlike the above work, here we target a different task: detecting previously fact-checked claims as opposed to check-worthiness prediction or fact-checking a claim. Moreover, while the above work was limited to the target context, here we also model the source context, which turns out to be much more important.

\begin{table*}[!tbh]
    \centering
    \small
    \resizebox{1.0\textwidth}{!}{% 
    \begin{tabular}{ll l p{0.3\textwidth}p{0.29\textwidth}}
        \toprule
        \bf Line No. & \bf Type & \multicolumn{2}{c}{\bf Input Claim} & \multicolumn{1}{c}{\bf Verified Claim} \\
        \midrule
        255 & \clean & D. Trump: & \emph{Hillary Clinton wanted the wall.} & Says Hillary Clinton ``wanted the wall.'' \\
        \midrule
        695 & \partof & C. Wallas: & \emph{And since then, as we all know, nine women have come forward and have said that you either groped them or kissed them without their consent.} & The stories from women saying he groped or forced himself on them ``largely have been debunked.'' \\
        \multicolumn{5}{c}{\bf \vdots} \\
        699 & \partof & D. Trump:  & \emph{Well, first of all, those stories have been largely debunked.} & The stories from women saying he groped or forced himself on them ``largely have been debunked.'' \\
        \midrule
        688 & \cleanhard & D. Trump: & \emph{She gave us ISIS as sure as you are sitting there.} & Hillary Clinton invented ISIS with her stupid policies. She is responsible for ISIS. \\
        \midrule
        605 & & D. Trump: & \emph{Now she wants to sign TransPacific Partnership.} & \\
        \multicolumn{5}{c}{\bf \vdots} \\
        607 & \contextdep & D. Trump: & \emph{She lied when she said she didn't call it the gold standard in one of the debates.} & Says Hillary Clinton called the TransPacific Partnership ``the gold standard. You called it the gold standard of trade deals. You said its the finest deal youve ever seen.'' \\

        \bottomrule
    \end{tabular}
    }
    \caption{Fragment from the 3rd US Presidential debate in 2016 showing the \textit{verified claims} chosen by \politifact~and the fine-grained category of the pair. Most input sentences have no \textit{verified claim}, e.g., see line 605.}
    \label{table:dataset-examples}
\end{table*}

\section{Dataset}
\label{sec:dataset}

Here, we focus on the problem of detecting previously fact-checked claims, using the task formulation and an adaptation of data from \cite{shaar-etal-2020-known}. They had two datasets: one on matching tweets against Snopes claims, and another one on matching claims in the context of a political debate to \politifact\ claims. Here, we focus on the latter,\footnote{\url{github.com/sshaar/That-is-a-Known-Lie}} 
and we perform a close analysis of the claims and what makes them easy/hard to match.

We experimented with their \politifact\  dataset, which targets claims related to US politics. After a US political debate, speech, or interview, fact-checking journalists from \politifact\ would select few claims made in the event and would verify them either from scratch or by linking them to a previously fact-checked claim. 
Each fact-checked claim has an associated article stating its degree of factuality along with an explanation of how the fact-checkers arrived at their verdict.
The dataset has two parts: (\emph{i})~verified claims \{normalized \vclaim, article {\em title}, and article {\em text}\}, (\emph{ii})~transcripts of the political events (e.g.,~debates). \citet{shaar-etal-2020-known} annotated the data by linking sentences from the transcript (\inputclaim) to one or more verified claims (out of 16,636 claims in PolitiFact).

To further analyze the dataset, we looked at the \inputclaim--\vclaim~pairs, and we manually categorized them into one of the following categories:
\begin{enumerate}
    \item \textbf{\clean:} A \clean~pair is a self-contained 
    \inputclaim~with a \vclaim~that directly verifies it (see line 255 in Table~\ref{table:dataset-examples} as an example). 
    \item \textbf{\cleanhard:} A \cleanhard~pair is a self-contained \inputclaim~with a \vclaim~that indirectly verifies it (see line 688 in Table~\ref{table:dataset-examples}).
    \item \textbf{\partof:} A \partof~ pair's \inputclaim~is not self-contained and requires the addition of other sentences from the transcript to fully form a single claim.
    \item \textbf{\contextdep:} A \contextdep~pair is similar to \clean~and \cleanhard, but the \inputclaim~is not self-contained and needs co-reference.
\end{enumerate}

The above categories include all types of pairs we have seen. Moreover, since the dataset is constructed from speeches, debates, and interviews, the structure of the \inputclaim--\vclaim~pairs differs. For example, in debates, we see more \partof\ examples, as there are multiple question--answer claim pairs, as well as back-and-forth arguments splitting the claims into multiple sentences.

The annotations were performed by three  annotators who are experts in fact-checking (and co-authors of this paper), using the above definitions for the categories. We consolidated their annotations using majority voting, and they had a consolidation discussion for cases with no majority. The Fleiss Kappa inter-annotator agreement was 0.5, which corresponds to moderate agreement, which is reasonable for such a complex annotation task.

Table \ref{table:dataset-examples} shows examples of \inputclaim--\vclaim~pairs that illustrate the four categories. We can see that the task goes beyond simple textual similarity and natural language inference, as the examples in lines 607 and 695--699 show. Moreover, matching \contextdep~pairs (lines 605--607) requires understanding the \inputclaim's local context, while matching \cleanhard~pairs (line 688) requires analysis of the overall global context of the \vclaim.

Finally, we should note while annotating the data into the above four categories, we found out that a small number of \inputclaim--\vclaim~pairs in \cite{shaar-etal-2020-known} were false matches (which happened, as they did the matching automatically, without manually double-checking every single example).
We removed these pairs, and thus our reported number of pairs is slightly lower than theirs.

\begin{table}[t]
    \scalebox{0.8}{
    \begin{tabular}{lrr}
        \toprule
        \emph{\inputclaim}--\emph{\vclaim}~pairs & \multicolumn{2}{c}{695} \\ \midrule
        -- \clean & 291 & 42\%\\
        -- \cleanhard & 210 & 30\%\\
        -- \partof & 68 & 10\%\\
        -- \contextdep & 126 & 18\%\\\midrule
        Total \# of verified claims (to match against) & \multicolumn{2}{c}{16,636} \\
        \bottomrule
    \end{tabular}
    }
    \caption{\textbf{Statistics about our dataset:} total number of \emph{\inputclaim}--\emph{\vclaim}~pairs and of \emph{VerClaims} in \politifact\ to match an \emph{InputClaim} against. }
  \label{table:datasets:statistics}
\end{table}

Table \ref{table:datasets:statistics} shows statistics about the distribution of the four categories of claims in our dataset. We can see that \clean~and \cleanhard~are the most frequent categories, while \partof~is the least frequent one.

We further observed that \citet{shaar-etal-2020-known} dealt with each \inputclaim~independently, i.e.,~at the sentence level. This is problematic because for \partof\ claims we could end up splitting them and putting them in different sets: one in training, and one in testing. Moreover, splitting the dataset in this way means that the examples for a given topic can split between training and test, and thus information can leak, e.g.,~a claim can be repeated. Therefore, we considered new splits for the data:
\begin{itemize}
    \item \emph{Debate-Level Chrono}: We split the data chronologically. We use the first 50 debates for training, and the last 20 for testing. Specifically, we have 554 pairs for training, and 141 pairs for testing. This is a more realistic scenario, where we would only have access to earlier debates, and we can use them to make decisions about claims made in future debates. The complexity of this setting is also reflected in the MAP score as shown in Table~\ref{table:baselines-results}. We see that this score is lower than the best model in previous work (last row). This is because this setting is complex as we use a model trained on debates and speeches from 2012-2018, and we test on debates from 2019. Across those different time frames, different politicians discuss different topics.
    \item \emph{Debate-Level Semi-Chrono}: We split the data per year, e.g., for year 2018, we divide the transcripts into train and test with 80/20 splits, and then we train and evaluate using the same reranking model. In Table~\ref{table:baselines-results}, we can see an improvement with this setting compared to the \emph{Debate-Level Chrono} setting. This might be because the same politicians discuss the same/similar issues throughout the same year. 
    \item \emph{Debate-Level Random}: We randomly choose 80\% of the debates for training and the remaining ones for testing. This is a comparatively easier setting as the data is randomly distributed in training and testing. This is also reflected in the results in Table~\ref{table:baselines-results}. The reason could be that politicians repeat themselves a lot, especially in two consecutive political events, and the random split can lead to having two similar debates/speeches in two splits. 
    \item \emph{Sentence Level Random}: This is the setting in \citep{shaar-etal-2020-known}, where \emph{sentences} from the debates are randomly divided into train and test in a proportion of 80:20. This is the most unrealistic split.
\end{itemize}

\begin{table}[t]
    \centering
    \resizebox{0.95\linewidth}{!}{%  
    \begin{tabular}{l r}
        \toprule
        \bf Split & \bf MAP  \\
        \midrule
        {\bf Debate-Level -- Chrono} & \bf 0.429\\
        \midrule 
        Debate-Level -- Semi-chrono & 0.539 \\%0.548 \\
        Debate-Level -- Random & 0.590 \\%0.621  \\
        Sentence-Level -- Random \citep{shaar-etal-2020-known} & 0.602 \\% 0.608 \\
        \bottomrule
    \end{tabular}
    }
    \caption{MAP scores of the reranker models when using four different splits representing different scenarios. We use \emph{Debate-Level -- Chrono} for our experiments.}
  \label{table:baselines-results}
\end{table}

In our experiments, we chose to use the most realistic, but also the hardest setup: \emph{Debate-Level Chrono}. As a result, our MAP score, when experimenting with the state-of-the-art model of \citep{shaar-etal-2020-known}, decreases from 0.602 to 0.429.

\section{Experimental Setup}
\label{sec:experimental_setup}

Below, we first introduce the experimental setup for our baseline, and then we describe our proposed model that takes the context of the input claim into account, both on the source and on the target side.

\subsection{Baseline}
\label{sec:baseline}

From our analysis of the dataset (described in Section \ref{sec:dataset}), we conclude that (\emph{i})~we need to resolve the references in the \inputclaim, (\emph{ii})~to capture the local context of the \inputclaim,~and (\emph{iii})~to encapsulate the global context of the \vclaim.

For our baseline, we use the setup of the state-of-the-art model of \citet{shaar-etal-2020-known}. We use the claim as a query against the full text of the documents using BM25. 
We then train a reranker on the top-100 BM25 results using rankSVM~\cite{herbrich1999support} with an RBF kernel. 

The reranker uses nine similarity measures that compare the 
\inputclaim~to the \vclaim, as well as the respective reciprocal ranks. In particular, we compute the BM25 score for \inputclaim~vs. \vclaim, {\em title}, {\em text}, \vclaim+{\em title}+{\em text}. We also compute the cosine using sentence-BERT embeddings for \inputclaim~vs. \vclaim, {\em title}, and the top-4 sentences from {\em text}.
Using these scores, we create a vector representation of the \inputclaim--\vclaim~pair with dimensionality $\mathbb{R}^{18}$. We then scale the vectors of all \inputclaim--\vclaim~pairs in $[-1; 1]$ and we train a rankSVM with default values of the hyper-parameters:
$KernelDegree = 3$, $\gamma = 1/num\_features$, and $\epsilon = 0.001$.

\subsection{Proposed Model}

As shown in Figure \ref{fig:previously_fack_check_pipeline}, our model uses co-reference resolution on the source and on the target side, the local context (i.e.,~the neighboring sentences), and the global context (using Transformer-XH) as discussed below. It is still a pairwise reranker, but with a richer context representation.

\subsubsection{Co-reference Resolution}
We manually inspected the training transcripts and the associated verified claims, and we realized that there were many co-reference dependencies, resolving which could potentially help to obtain more representative textual and contextual similarity scores. 
As for the verified claims, we noticed that not all {\vclaim}s were self-contained, and that some understanding of the context was needed\footnote{For example, who is speaking or what is being discussed.} of the article's {\em text} that explains the verdict provided by the \politifact\  journalists. Therefore, our hypothesis was that resolving such co-references could improve the downstream matching scores.

For the same reason, we also performed co-reference resolution on the \politifact\ articles when they were used to compute the BM25 scores.

We experimented with various co-reference resolution tools including \textbf{NeuralCoref},\footnote{\url{github.com/huggingface/neuralcoref}} 
\textbf{e2e-coref},\footnote{\url{github.com/kentonl/e2e-coref}} and \textbf{SpanBERT},\footnote{\url{github.com/facebookresearch/SpanBERT}} and we found that \textbf{NeuralCoref} was best on the input transcripts, while \textbf{e2e-coref} was best on the articles about the target {\vclaim}s. Hence, in the rest of our experiments below, we show results using \textbf{NeuralCoref} for the source side, and using \textbf{e2e-coref} for the target side.

We resolved the co-reference in the \inputclaim~ by performing co-reference resolution on the entire input transcript (as was suggested in the literature); we will refer to this as \emph{\textbf{src-coref}}. 
As for the verified claims, we aimed to resolve the co-references both in the \vclaim~and in the \emph{text} of the \politifact\ articles. We also aimed to ensure that the dependencies from the \emph{text} can be used for the \vclaim. Therefore, we concatenated both the text and \vclaim~(in the same order), and we applied the co-reference model on the concatenated text. We chose this order of concatenation because the published \emph{text} reserves the last paragraph to rephrase the \vclaim~and to provide a summary of the justification; hence, there is a higher probability to resolve the co-references correctly.

\subsubsection{Local Context}
\label{ssec:local_context}

Resolving the pronominal co-references allows us to obtain the correct objects and the names the \textit{InputClaim} refers to. However, in the process of analyzing the dataset, we noticed that different {\vclaim}s, although having similar structure, could talk about different things, depending on the article text and also on the surrounding context. Therefore, it is important to understand the context of an \inputclaim.
In particular, we achieve this by performing a feature-level concatenation of the neighboring sentences in the transcript, i.e.,~we take the eighteen features ($\mathbb{R}^{18}$, as discussed in Section~\ref{sec:baseline} above) for the neighboring sentences, and we concatenate them to the similarity score for the \inputclaim. We then use the resulting representation as a feature vector to be fed into our reranker. For example, if we take three sentences before the \inputclaim~and one sentence after it, we denote this as \featurecontext(3, 1).

Let $S_{i}$ be our \inputclaim, which is the $i$'th sentence in the transcript. We compute the similarity measures and the reciprocal rank (as described in Section~\ref{sec:baseline}) to obtain the vector representation $S_{i,v}$ for $S_{i}$. 
With $k=3$ previous and $l=1$ following neighbouring sentences our final feature vector is 
\begin{multline}
\footnotesize
\label{eq:feature-context}
FC(k=3, l=1)=S_{i-3,v}\concat S_{i-2,v} \\\concat S_{i-1,v}
\concat S_{i,v} \concat S_{i+1,v}  
\end{multline}
where
$\concat$ represents concatenation.

Note that after the concatenation, the resulting dimensionality of the feature vector for \textit{\textbf{\featurecontext(3, 1)}} is $18 \times (3+1+1) = 90$. 

\begin{table*}[tbh]
    \small
    \centering
    \setlength{\tabcolsep}{4pt}
    \scalebox{1.0}{
    \begin{tabular}{c p{0.4\textwidth} c c*{4}{c} }
        \toprule
        \bf Line No. & \bf Model & \bf Overall & \bf \clean & \bf \cleanhard & \bf \partof & \bf \contextdep \\
        \midrule
        1 & Baseline & 0.429 & 0.661 & 0.365 & 0.161 & 0.375 \\
        \midrule
        \multicolumn{7}{c}{\bf Source-Side Experiments: Co-reference Resolution, Local Context}\\
        \midrule
		2 & \featurecontext(3, 1) & 0.513 & 0.690 & \bf 0.485 & 0.305 & 0.448\\
		3 & src-coref & 0.479 & 0.667 & 0.408 & 0.286 & 0.429\\
		4 & src-coref + \featurecontext(3, 1) & \bf 0.532 & 0.695 &	0.452 &	\bf 0.385 &	\bf 0.485\\
		\midrule
		\multicolumn{7}{c}{\bf Target-Side Experiments: Co-reference Resolution, Global Context}\\
		\midrule
		5 & \textit{Transformer-XH} & 0.468 & 0.680 & 0.441 & 0.226 & 0.384 \\
		6 & tgt-coref & 0.443 & 0.673 & 0.422 & 0.182 & 0.339 \\
		7 & tgt-coref + \textit{Transformer-XH} & 0.458 & 0.702 & 0.444 & 0.161 & 0.357\\
		\midrule
		\multicolumn{7}{c}{\bf Source+Target-Side Experiments: Co-reference Resolution, Local Context, Global Context}\\
		\midrule
		8 & src-coref + tgt-coref & 0.487 & 0.672	& 0.440	& 0.291	& 0.411\\
        9 & All & 0.517 & \bf 0.749 & 0.389 & 0.321 & 0.464\\	
        \bottomrule
    \end{tabular}
    }
    \caption{MAP scores of the reranker models on the test set using the \textit{Debate-Level Chrono} split.}
  \label{table:results}
\end{table*}

\subsubsection{Global Context}
\label{ssec:global_context}
The similarity scores that leverage the local context in the textual content of the \inputclaim~and the \vclaim~are obtained using (\emph{i})~BM25, and (\emph{ii})~the cosine similarity between the Sentence-BERT embeddings of the \inputclaim~vs. the top-4 sentences of the \vclaim. This might miss relevant information further away from the \inputclaim~in the input document and further away from the \vclaim~in the document accompanying the \vclaim.
We refer to such scattered information as the \textbf{global context}. To capture it, we use Transformer-XH  \cite{zhao2019transformer}, which is pretrained on the FEVER (Fact Extraction and VERification) dataset to predict whether a given input claim is supported/refuted by a set of target sentences (from Wikipedia), represented as a graph, or there is no enough information. We used the model from \citep{zhao2019transformer}.
For a given \inputclaim, we generate a graph for each of the top-100 {\vclaim}s retrieved using BM25 and the normalized claim, the {\em title}, and the top-3 sentences from the {\em text} as nodes.
Using the \emph{\textbf{Transformer-XH}} model on the graph, we obtain three additional scores that correspond to the posterior probability that \vclaim~supports or refutes the \inputclaim, or there is no enough information.

\subsection{Hyper-Parameter Values}
For the baseline, we use the best values of the hyper-parameters as found in \cite{shaar-etal-2020-known}.
For our context-aware models, we select the values of the hyper-parameters by splitting the training dataset into train-train (debates from 2012-2017) and train-dev (debates from 2018), then we train on the former, and we test on the latter.

\subsection{Evaluation Measures}

As we have a ranking task, we use mean average precision (MAP) for evaluation. It is a suitable measure as some {\inputclaim}s are paired with more than one \vclaim. This is why we opted for not using mean reciprocal rank (MRR), which would only pay attention to the highest-ranked match.

\section{Results}
\label{sec:results}

Below, we described the results for our source-side and target-side context modeling experiments.

\subsection{Source-Side Experiments}

For the source side experiments, we used co-reference resolution on transcripts and variations of the local context by varying $k$ and $l$ in Eq.~\ref{eq:feature-context}.

When we inspected the transcripts, we found that co-references tended to be resolved by a few sentences before the \inputclaim; therefore, we tried \featurecontext(1, 1), \featurecontext(3, 1), \featurecontext(3, 3), and \featurecontext(5, 1).
We obtained the best results on cross-validation using \featurecontext(3, 1), which we use below.
As shown in Table~\ref{table:results}, the local context (line 2) improves over the baseline (line 1) by eight MAP points absolute.

We experimented using co-reference resolution with \textbf{NeuralCoref}. 
This yielded a sizable improvement over the baseline as shown in line 3 in Table~\ref{table:results}, especially for \partof~and \contextdep\ pairs, as they have many co-references, which can make it hard for the model the understand the \inputclaim. After combining the two methods, i.e.,~\textit{src-coref} and \featurecontext(3,1) (see line 4), we achieved the highest MAP score of 0.532.
We always see an improvement for the \clean~category as the resolved \inputclaim~can match the article text better.

\subsection{Target-Side Experiments}

For the target-side experiments, we tried using co-reference resolution (on the source and on the target side) for the \vclaim~and the fact-checking article, as well as modeling the global context with \textit{Transformer-XH}. 
Compared to the baseline, we see on line 5 of Table~\ref{table:results} a sizable improvement from 0.365 to 0.441 MAP points for \cleanhard. 

This is expected as the pair does not exhibit much semantic similarity, and we need to build our own understanding of the {\em text} of the \vclaim~in order to capture the contextual similarity in the pair. We also experimented with co-reference resolution on the \vclaim~and the \emph{text} of the \emph{\vclaim} and also see some improvement. Combining \textit{tgt-coref} and \textit{Transformer-XH} (line 7) improved the performance over \textit{tgt-coref} alone, but it is worse than \textit{Transformer-XH} alone. The combination outperforms other target-side experiments for \textit{clean}.  

\subsection{Source-Side \& Target-Side Experiments}

Eventually, we experimented with modeling the context both on the source and on the target side. Line 8 in Table~\ref{table:results} shows the evaluation results when we use co-reference resolution both on the source and on the target side. We can see that this yields a higher overall MAP score of 0.487, compared to using source-side (MAP of 0.479; line 3) or target-side context only (MAP of 0.443; line 6). Moreover, co-reference resolution on both sides helps for \cleanhard~and \partof (compared to using co-reference on one side only) as they require better local and global context, respectively.

We further tried putting it all together, and the result is shown in line 9.\footnote{Note that in this result we did not use target-side co-reference, as adding it yielded somewhat worse results. It seems to interact badly with Transformer-XH, which can also be seen by comparing lines 5 and 7.} While this yielded better results for \emph{clean}, it was slightly worse compared to the source-side context modeling combination in line 4. This is probably due to the source-side context models being generally stronger than the target-side ones (compare lines 2--3 to lines 5--6).

We can conclude that modeling the context on the source side is much more important than on the target side. This is expected for political debates, which are conversational in nature. In contrast, the target side is a well-written journalistic article, where sentences are much more self-contained. Thus, features from the source side (i.e.,~from the debate) are more useful as can be seen in Table~\ref{table:results}.

\subsection{Discussion}

As mentioned above, our baseline is a reimplementation of the best system of \citet{shaar-etal-2020-known}, and our context modeling extensions add additional components on top of it. Note, however, that our experimental results are not directly comparable to their published ones, as we use a more realistic and also a much harder setup, where the data is split by entire debates and also chronologically, following the \emph{Debate-Level Chrono} data split, as we discussed in Section~\ref{sec:dataset}, i.e.,~training on the data from 2012 to 2018 and testing on 2019 (while they split all debates into sentences and randomly distribute them to training/testing). However, we do have comparison to their approach, as we ran their model on our data split, which is our baseline, as shown on line 1 of Table~\ref{table:results}.

\section{Conclusion and Future Work}
\label{sec:conclusion}

We have presented our work on the important but under-studied problem of detecting previously fact-checked claims in political debates and speeches. We studied the impact of modeling the context: both on the source side, i.e.,~in the debate, as well as on the target side, i.e.,~in the fact-checking document that explains how human fact-checkers have arrived at their decision about the factuality of the claim. In particular, we modeled the local context, the global context, and we further used co-reference resolution and multi-hop reasoning over the target text using Transformer-XH. The experimental results have shown that each of these components represents a valuable information source, but modeling the source-side context is more important, and can yield 10+ points of absolute improvement over a context-free state-of-the-art baseline.

In future work, we want to try other multi-hop reasoning frameworks for context modeling. We also plan to experiment with other kinds of conversations, e.g.,~in community forums and in social media, including for other languages.

\section*{Acknowledgments}
This research is part of the Tanbih mega-project, developed at the Qatar Computing Research Institute, HBKU, which aims to limit the impact of ``fake news,'' propaganda, and media bias by making users aware of what they are reading, thus promoting media literacy and critical thinking.

\section*{Ethics and Broader Impact}

\paragraph{Biases}

We note that there might be some biases in the data we use, as well as in some manual judgments for claim matching. There could be also biases in the data selection and the fact-checking process of the human fact-checkers, which are beyond our control. Finally, there are known biases in the large-scale pre-trained transformer models that we experiment with.

\paragraph{Intended Use and Misuse Potential}

Our models can make it possible to put politicians on the spot in real time, e.g., during an interview or a political debate, by providing journalists with tools to do trustable fact-checking in real time. They can also save a lot of time to fact-checkers for unnecessary double-checking something that was already fact-checked. However, these models could also be misused by malicious actors. We, therefore, ask researchers to exercise caution.

\paragraph{Environmental Impact}
We would also like to warn that the use of large-scale Transformers requires a lot of computations and the use of GPUs/TPUs for training, which contributes to global warming \cite{strubell-etal-2019-energy}. This is a bit less of an issue in our case, as we do not train such models from scratch; rather, we fine-tune them on relatively small datasets. Moreover, running on a CPU for inference, once the model has been fine-tuned, is perfectly feasible, and CPUs contribute much less to global warming.

\bibliographystyle{acl_natbib}
\bibliography{bib/anthology,bib/acl2021}

\begin{thebibliography}{61}
\expandafter\ifx\csname natexlab\endcsname\relax\def\natexlab#1{#1}\fi

\bibitem[{Atanasova et~al.(2018)Atanasova, M\`{a}rquez, Barr\'{o}n-Cede\~{n}o,
  Elsayed, Suwaileh, Zaghouani, Kyuchukov, Da~San~Martino, and
  Nakov}]{clef2018checkthat:task1}
Pepa Atanasova, Llu\'{i}s M\`{a}rquez, Alberto Barr\'{o}n-Cede\~{n}o, Tamer
  Elsayed, Reem Suwaileh, Wajdi Zaghouani, Spas Kyuchukov, Giovanni
  Da~San~Martino, and Preslav Nakov. 2018.
\newblock Overview of the {CLEF-2018 CheckThat!} lab on automatic
  identification and verification of political claims, task 1:
  Check-worthiness.
\newblock In \emph{CLEF 2018 Working Notes. Working Notes of CLEF 2018 -
  Conference and Labs of the Evaluation Forum}, {CEUR} Workshop Proceedings,
  Avignon, France.

\bibitem[{Atanasova et~al.(2019{\natexlab{a}})Atanasova, Nakov, Karadzhov,
  Mohtarami, and Martino}]{clef-checkthat-T1:2019}
Pepa Atanasova, Preslav Nakov, Georgi Karadzhov, Mitra Mohtarami, and Giovanni
  Da~San Martino. 2019{\natexlab{a}}.
\newblock Overview of the {CLEF-2019 CheckThat! Lab on Automatic Identification
  and Verification of Claims. Task 1: Check-Worthiness}.
\newblock In \emph{CLEF 2019 Working Notes}, Lugano, Switzerland.

\bibitem[{Atanasova et~al.(2019{\natexlab{b}})Atanasova, Nakov, M{\`a}rquez,
  Barr{\'o}n-Cede{\~n}o, Karadzhov, Mihaylova, Mohtarami, and
  Glass}]{atanasova2019automatic}
Pepa Atanasova, Preslav Nakov, Llu{\'\i}s M{\`a}rquez, Alberto
  Barr{\'o}n-Cede{\~n}o, Georgi Karadzhov, Tsvetomila Mihaylova, Mitra
  Mohtarami, and James Glass. 2019{\natexlab{b}}.
\newblock Automatic fact-checking using context and discourse information.
\newblock \emph{Journal of Data and Information Quality (JDIQ)}, 11(3):1--27.

\bibitem[{Augenstein et~al.(2019)Augenstein, Lioma, Wang, Chaves~Lima, Hansen,
  Hansen, and Simonsen}]{augenstein-etal-2019-multifc}
Isabelle Augenstein, Christina Lioma, Dongsheng Wang, Lucas Chaves~Lima, Casper
  Hansen, Christian Hansen, and Jakob~Grue Simonsen. 2019.
\newblock \href {https://doi.org/10.18653/v1/D19-1475} {{M}ulti{FC}: A
  real-world multi-domain dataset for evidence-based fact checking of claims}.
\newblock In \emph{Proceedings of the 2019 Conference on Empirical Methods in
  Natural Language Processing and the 9th International Joint Conference on
  Natural Language Processing}, EMNLP-IJCNLP~'19', pages 4685--4697, Hong Kong,
  China.

\bibitem[{Barr\'{o}n-Cede{\~n}o et~al.(2020)Barr\'{o}n-Cede{\~n}o, Elsayed,
  Nakov, {Da San Martino}, Hasanain, Suwaileh, Haouari, Babulkov, Hamdan,
  Nikolov, Shaar, and {Sheikh Ali}}]{clef-checkthat-lncs:2020}
Alberto Barr\'{o}n-Cede{\~n}o, Tamer Elsayed, Preslav Nakov, Giovanni {Da San
  Martino}, Maram Hasanain, Reem Suwaileh, Fatima Haouari, Nikolay Babulkov,
  Bayan Hamdan, Alex Nikolov, Shaden Shaar, and Zien {Sheikh Ali}. 2020.
\newblock {Overview of CheckThat! 2020}: Automatic identification and
  verification of claims in social media.
\newblock In \emph{Experimental IR Meets Multilinguality, Multimodality, and
  Interaction Proceedings of the Eleventh International Conference of the CLEF
  Association (CLEF 2020)}, LNCS (12260).

\bibitem[{Barr{\'{o}}n{-}Cede{\~{n}}o et~al.(2020)Barr{\'{o}}n{-}Cede{\~{n}}o,
  Elsayed, Nakov, Martino, Hasanain, Suwaileh, and
  Haouari}]{CheckThat:ECIR2020}
Alberto Barr{\'{o}}n{-}Cede{\~{n}}o, Tamer Elsayed, Preslav Nakov, Giovanni
  Da~San Martino, Maram Hasanain, Reem Suwaileh, and Fatima Haouari. 2020.
\newblock {CheckThat! at CLEF} 2020: Enabling the automatic identification and
  verification of claims in social media.
\newblock In \emph{Proceedings of the 42nd European Conference on Information
  Retrieval}, ECIR~'20, pages 499--507, Lisbon, Portugal.

\bibitem[{Barr\'{o}n-Cede{\~n}o et~al.(2018)Barr\'{o}n-Cede{\~n}o, Elsayed,
  Suwaileh, Marquez, Atanasova, Zaghouani, Kyuchukov, Da~San~Martino, and
  Nakov}]{clef-checkthat-T2:2018}
Alberto Barr\'{o}n-Cede{\~n}o, Tamer Elsayed, Reem Suwaileh, Lluis Marquez,
  Pepa Atanasova, Wajdi Zaghouani, Spas Kyuchukov, Giovanni Da~San~Martino, and
  Preslav Nakov. 2018.
\newblock Overview of the {CLEF-2018 CheckThat!} lab on automatic
  identification and verification of political claims. {T}ask 2: Factuality.
\newblock In \emph{CLEF 2018 Working Notes. Working Notes of CLEF 2018 -
  Conference and Labs of the Evaluation Forum}, {CEUR} Workshop Proceedings,
  Avignon, France.

\bibitem[{Bouziane et~al.(2020)Bouziane, Perrin, Cluzeau, Mardas, and
  Sadeq}]{clef-checkthat-Bouziane:2020}
Mostafa Bouziane, Hugo Perrin, Aur\`elien Cluzeau, Julien Mardas, and Amine
  Sadeq. 2020.
\newblock {Buster.AI} at {CheckThat!} 2020: {I}nsights and recommendations to
  improve fact-checking.
\newblock In \emph{Working Notes of CLEF 2020---Conference and Labs of the
  Evaluation Forum}, CLEF~'2020, Thessaloniki, Greece.

\bibitem[{Chen et~al.(2017)Chen, Zhu, Ling, Wei, Jiang, and
  Inkpen}]{chen2016enhanced}
Qian Chen, Xiaodan Zhu, Zhen-Hua Ling, Si~Wei, Hui Jiang, and Diana Inkpen.
  2017.
\newblock Enhanced {LSTM} for natural language inference.
\newblock In \emph{Proceedings of the 55th Annual Meeting of the Association
  for Computational Linguistics}, ACL~'17, pages 1657--1668, Vancouver, Canada.

\bibitem[{Chernyavskiy et~al.(2022)Chernyavskiy, Ilvovsky, Kalinin, and
  Nakov}]{NAACL:2022:Batch:loss}
Anton Chernyavskiy, Dmitry Ilvovsky, Pavel Kalinin, and Preslav Nakov. 2022.
\newblock Batch-softmax contrastive loss for pairwise sentence scoring tasks.
\newblock In \emph{Proceedings of the Annual Conference of the North American
  Chapter of the Association for Computational Linguistics: Human Language
  Technologies}, NAACL-HLT~'22, Seattle, Washington, USA.

\bibitem[{Chernyavskiy et~al.(2021{\natexlab{a}})Chernyavskiy, Ilvovsky, and
  Nakov}]{clef-checkthat:2021:task3:Chernyavskiy2021}
Anton Chernyavskiy, Dmitry Ilvovsky, and Preslav Nakov. 2021{\natexlab{a}}.
\newblock Aschern at {CLEF CheckThat!} 2021: Lambda-calculus of fact-checked
  claims.
\newblock In \emph{{CLEF} 2021 Working Notes. {W}orking Notes of {CLEF}
  2021--Conference and Labs of the Evaluation Forum}, Bucharest, Romania
  (online).

\bibitem[{Chernyavskiy et~al.(2021{\natexlab{b}})Chernyavskiy, Ilvovsky, and
  Nakov}]{WhatTheWikiFact}
Anton Chernyavskiy, Dmitry Ilvovsky, and Preslav Nakov. 2021{\natexlab{b}}.
\newblock \href {https://doi.org/10.1145/3459637.3481987} {{WhatTheWikiFact}:
  Fact-checking claims against {W}ikipedia}.
\newblock In \emph{Proceedings of the 30th ACM International Conference on
  Information and Knowledge Management}, CIKM~'21, pages 4690--4695.

\bibitem[{Gencheva et~al.(2017)Gencheva, Nakov, M{\`a}rquez,
  Barr{\'o}n-Cede{\~n}o, and Koychev}]{gencheva2017context}
Pepa Gencheva, Preslav Nakov, Llu{\'\i}s M{\`a}rquez, Alberto
  Barr{\'o}n-Cede{\~n}o, and Ivan Koychev. 2017.
\newblock \href {https://doi.org/10.26615/978-954-452-049-6_037} {A
  context-aware approach for detecting worth-checking claims in political
  debates}.
\newblock In \emph{Proceedings of the International Conference Recent Advances
  in Natural Language Processing}, RANLP~'17, pages 267--276, Varna, Bulgaria.

\bibitem[{Hanselowski et~al.(2018)Hanselowski, Zhang, Li, Sorokin, Schiller,
  Schulz, and Gurevych}]{hanselowski2018ukp}
Andreas Hanselowski, Hao Zhang, Zile Li, Daniil Sorokin, Benjamin Schiller,
  Claudia Schulz, and Iryna Gurevych. 2018.
\newblock \href {https://doi.org/10.18653/v1/W18-5516} {{UKP}-{A}thene:
  Multi-sentence textual entailment for claim verification}.
\newblock In \emph{Proceedings of the First Workshop on Fact Extraction and
  {VER}ification}, FEVER~'18, pages 103--108, Brussels, Belgium.

\bibitem[{Hasanain et~al.(2020)Hasanain, Haouari, Suwaileh, Ali, Hamdan,
  Elsayed, Barr\'{o}n-Cede{\~n}o, {Da San Martino}, and
  Nakov}]{clef-checkthat-ar:2020}
Maram Hasanain, Fatima Haouari, Reem Suwaileh, {Zien Sheikh} Ali, Bayan Hamdan,
  Tamer Elsayed, Alberto Barr\'{o}n-Cede{\~n}o, Giovanni {Da San Martino}, and
  Preslav Nakov. 2020.
\newblock Overview of {CheckThat!} 2020 {A}rabic: Automatic identification and
  verification of claims in social media.
\newblock In \emph{Working Notes of CLEF 2020---Conference and Labs of the
  Evaluation Forum}, CLEF~'2020, Thessaloniki, Greece.

\bibitem[{Hassan et~al.(2017)Hassan, Zhang, Arslan, Caraballo, Jimenez,
  Gawsane, Hasan, Joseph, Kulkarni, Nayak, Sable, Li, and
  Tremayne}]{Hassan:2017:CFE:3137765.3137815}
Naeemul Hassan, Gensheng Zhang, Fatma Arslan, Josue Caraballo, Damian Jimenez,
  Siddhant Gawsane, Shohedul Hasan, Minumol Joseph, Aaditya Kulkarni,
  Anil~Kumar Nayak, Vikas Sable, Chengkai Li, and Mark Tremayne. 2017.
\newblock {ClaimBuster}: The first-ever end-to-end fact-checking system.
\newblock \emph{Proc. VLDB Endow.}, 10(12):1945--1948.

\bibitem[{Herbrich et~al.(1999)Herbrich, Graepel, and
  Obermayer}]{herbrich1999support}
Ralf Herbrich, Thore Graepel, and Klaus Obermayer. 1999.
\newblock \href {https://doi.org/10.1049/cp:19991091} {Support vector learning
  for ordinal regression}.
\newblock In \emph{Proceedings of the 1999 Ninth International Conference on
  Artificial Neural Networks}, ICANN~'99, pages 97--102, Edinburgh, UK.

\bibitem[{Hidey and Diab(2018)}]{hidey-diab-2018-team}
Christopher Hidey and Mona Diab. 2018.
\newblock \href {https://doi.org/10.18653/v1/W18-5525} {Team {SWEEP}er: Joint
  sentence extraction and fact checking with pointer networks}.
\newblock In \emph{Proceedings of the First Workshop on Fact Extraction and
  {VER}ification ({FEVER})}, pages 150--155, Brussels, Belgium.

\bibitem[{Lazer et~al.(2018)Lazer, Baum, Benkler, Berinsky, Greenhill, Menczer,
  Metzger, Nyhan, Pennycook, Rothschild, Schudson, Sloman, Sunstein, Thorson,
  Watts, and Zittrain}]{Lazer1094}
David~M.J. Lazer, Matthew~A. Baum, Yochai Benkler, Adam~J. Berinsky, Kelly~M.
  Greenhill, Filippo Menczer, Miriam~J. Metzger, Brendan Nyhan, Gordon
  Pennycook, David Rothschild, Michael Schudson, Steven~A. Sloman, Cass~R.
  Sunstein, Emily~A. Thorson, Duncan~J. Watts, and Jonathan~L. Zittrain. 2018.
\newblock The science of fake news.
\newblock \emph{Science}, 359(6380):1094--1096.

\bibitem[{Li et~al.(2016)Li, Gao, Meng, Li, Su, Zhao, Fan, and
  Han}]{Li:2016:STD:2897350.2897352}
Yaliang Li, Jing Gao, Chuishi Meng, Qi~Li, Lu~Su, Bo~Zhao, Wei Fan, and Jiawei
  Han. 2016.
\newblock \href {https://doi.org/10.1145/2897350.2897352} {A survey on truth
  discovery}.
\newblock \emph{{ACM SIGKDD} Explorations Newsletter}, 17(2):1--16.

\bibitem[{Liu et~al.(2020)Liu, Xiong, Sun, and Liu}]{liu-etal-2020-fine}
Zhenghao Liu, Chenyan Xiong, Maosong Sun, and Zhiyuan Liu. 2020.
\newblock \href {https://doi.org/10.18653/v1/2020.acl-main.655} {Fine-grained
  fact verification with kernel graph attention network}.
\newblock In \emph{Proceedings of the 58th Annual Meeting of the Association
  for Computational Linguistics}, pages 7342--7351, Online.

\bibitem[{Luken et~al.(2018)Luken, Jiang, and de~Marneffe}]{luken2018qed}
Jackson Luken, Nanjiang Jiang, and Marie-Catherine de~Marneffe. 2018.
\newblock \href {https://doi.org/10.18653/v1/W18-5526} {{QED}: A fact
  verification system for the {FEVER} shared task}.
\newblock In \emph{Proceedings of the First Workshop on Fact Extraction and
  {VER}ification ({FEVER})}, pages 156--160, Brussels, Belgium.

\bibitem[{Mihaylova et~al.(2021)Mihaylova, Borisova, Chemishanov, Hadzhitsanev,
  Hardalov, and Nakov}]{clef-checkthat:2021:task2:DIPS}
Simona Mihaylova, Iva Borisova, Dzhovani Chemishanov, Preslav Hadzhitsanev,
  Momchil Hardalov, and Preslav Nakov. 2021.
\newblock {DIPS at CheckThat!} 2021: Verified claim retrieval.
\newblock In \emph{{CLEF} 2021 Working Notes. {W}orking Notes of {CLEF}
  2021--Conference and Labs of the Evaluation Forum}, Bucharest, Romania
  (online).

\bibitem[{Mihaylova et~al.(2019)Mihaylova, Karadzhov, Atanasova, Baly,
  Mohtarami, and Nakov}]{mihaylova-etal-2019-semeval}
Tsvetomila Mihaylova, Georgi Karadzhov, Pepa Atanasova, Ramy Baly, Mitra
  Mohtarami, and Preslav Nakov. 2019.
\newblock \href {https://doi.org/10.18653/v1/S19-2149} {{S}em{E}val-2019 task
  8: Fact checking in community question answering forums}.
\newblock In \emph{Proceedings of the 13th International Workshop on Semantic
  Evaluation}, pages 860--869, Minneapolis, Minnesota, USA.

\bibitem[{Nakov et~al.(2022{\natexlab{a}})Nakov, Barr\'{o}n-Cede\~{n}o,
  Da~San~Martino, Alam, M\'{\i}guez, Caselli, Kutlu, Zaghouani, Li, Shaar,
  Shahi, Mubarak, Nikolov, Kartal, and Beltr\'{a}n}]{clef-checkthat:2022:task1}
Preslav Nakov, Alberto Barr\'{o}n-Cede\~{n}o, Giovanni Da~San~Martino, Firoj
  Alam, Rub\'{e}n M\'{\i}guez, Tommaso Caselli, Mucahid Kutlu, Wajdi Zaghouani,
  Chengkai Li, Shaden Shaar, Gautam~Kishore Shahi, Hamdy Mubarak, Alex Nikolov,
  Yavuz~Selim Kartal, and Javier Beltr\'{a}n. 2022{\natexlab{a}}.
\newblock Overview of the {CLEF}-2022 {CheckThat}! lab task 1 on identifying
  relevant claims in tweets.
\newblock In \emph{Working Notes of CLEF 2022---Conference and Labs of the
  Evaluation Forum}, CLEF~'2022, Bologna, Italy.

\bibitem[{Nakov et~al.(2022{\natexlab{b}})Nakov, Barr\'{o}n-Cede\~{n}o,
  Da~San~Martino, Alam, Stru\ss{}, Mandl, M\'{\i}guez, Caselli, Kutlu,
  Zaghouani, Li, Shaar, Shahi, Mubarak, Nikolov, Babulkov, Kartal, and
  Beltr\'{a}n}]{CheckThat:ECIR2022}
Preslav Nakov, Alberto Barr\'{o}n-Cede\~{n}o, Giovanni Da~San~Martino, Firoj
  Alam, Julia~Maria Stru\ss{}, Thomas Mandl, Rub\'{e}n M\'{\i}guez, Tommaso
  Caselli, Mucahid Kutlu, Wajdi Zaghouani, Chengkai Li, Shaden Shaar,
  Gautam~Kishore Shahi, Hamdy Mubarak, Alex Nikolov, Nikolay Babulkov,
  Yavuz~Selim Kartal, and Javier Beltr\'{a}n. 2022{\natexlab{b}}.
\newblock \href {https://doi.org/10.1007/978-3-030-99739-7_52} {The {CLEF-2022
  CheckThat!} lab on fighting the {COVID-19} infodemic and fake news
  detection}.
\newblock In \emph{Proceedings of the 44th European Conference on IR Research:
  Advances in Information Retrieval}, ECIR~'22, pages 416--428, Stavanger,
  Norway.

\bibitem[{Nakov et~al.(2022{\natexlab{c}})Nakov, Barr\'{o}n-Cede\~{n}o,
  Da~San~Martino, Alam, Stru\ss{}, Mandl, M\'{\i}guez, Caselli, Kutlu,
  Zaghouani, Li, Shaar, Shahi, Mubarak, Nikolov, Babulkov, Kartal, and
  Beltr\'{a}n}]{clef-checkthat:2022:LNCS}
Preslav Nakov, Alberto Barr\'{o}n-Cede\~{n}o, Giovanni Da~San~Martino, Firoj
  Alam, Julia~Maria Stru\ss{}, Thomas Mandl, Rub\'{e}n M\'{\i}guez, Tommaso
  Caselli, Mucahid Kutlu, Wajdi Zaghouani, Chengkai Li, Shaden Shaar,
  Gautam~Kishore Shahi, Hamdy Mubarak, Alex Nikolov, Nikolay Babulkov,
  Yavuz~Selim Kartal, and Javier Beltr\'{a}n. 2022{\natexlab{c}}.
\newblock Overview of the {CLEF}-2022 {CheckThat}! lab on fighting the
  {COVID-19} infodemic and fake news detection.
\newblock In \emph{Proceedings of the 13th International Conference of the CLEF
  Association: Information Access Evaluation meets Multilinguality,
  Multimodality, and Visualization}, CLEF~'2022, Bologna, Italy.

\bibitem[{Nakov et~al.(2018)Nakov, Barr\'{o}n-Cede{\~n}o, Elsayed, Suwaileh,
  M\`{a}rquez, Zaghouani, Gencheva, Kyuchukov, and {Da San
  Martino}}]{clef2018checkthat}
Preslav Nakov, Alberto Barr\'{o}n-Cede{\~n}o, Tamer Elsayed, Reem Suwaileh,
  Llu\'{i}s M\`{a}rquez, Wajdi Zaghouani, Pepa Gencheva, Spas Kyuchukov, and
  Giovanni {Da San Martino}. 2018.
\newblock Overview of the {CLEF}-2018 lab on automatic identification and
  verification of claims in political debates.
\newblock In \emph{Working Notes of {CLEF} 2018 -- Conference and Labs of the
  Evaluation Forum}, CLEF~'18, Avignon, France.

\bibitem[{Nakov et~al.(2021{\natexlab{a}})Nakov, Corney, Hasanain, Alam,
  Elsayed, Barr{\'{o}}n{-}Cede{\~{n}}o, Papotti, Shaar, and
  Martino}]{Survey:2021:AI:Fact-Checkers}
Preslav Nakov, David Corney, Maram Hasanain, Firoj Alam, Tamer Elsayed, Alberto
  Barr{\'{o}}n{-}Cede{\~{n}}o, Paolo Papotti, Shaden Shaar, and Giovanni Da~San
  Martino. 2021{\natexlab{a}}.
\newblock Automated fact-checking for assisting human fact-checkers.
\newblock In \emph{Proceedings of the 30th International Joint Conference on
  Artificial Intelligence}, IJCAI~'21, pages 4551--4558.

\bibitem[{Nakov et~al.(2022{\natexlab{d}})Nakov, Da~San~Martino, Alam, Shaar,
  Mubarak, and Babulkov}]{clef-checkthat:2022:task2}
Preslav Nakov, Giovanni Da~San~Martino, Firoj Alam, Shaden Shaar, Hamdy
  Mubarak, and Nikolay Babulkov. 2022{\natexlab{d}}.
\newblock Overview of the {CLEF}-2022 {CheckThat}! lab task 2 on detecting
  previously fact-checked claims.
\newblock In \emph{Working Notes of CLEF 2022---Conference and Labs of the
  Evaluation Forum}, CLEF~'2022, Bologna, Italy.

\bibitem[{Nakov et~al.(2021{\natexlab{b}})Nakov, Da~San~Martino, Elsayed,
  Barr{\'o}n-Cedeno, M{\'\i}guez, Shaar, Alam, Haouari, Hasanain, Babulkov
  et~al.}]{nakov2021clef}
Preslav Nakov, Giovanni Da~San~Martino, Tamer Elsayed, Alberto
  Barr{\'o}n-Cedeno, Rub{\'e}n M{\'\i}guez, Shaden Shaar, Firoj Alam, Fatima
  Haouari, Maram Hasanain, Nikolay Babulkov, et~al. 2021{\natexlab{b}}.
\newblock The {CLEF-2021 CheckThat!} lab on detecting check-worthy claims,
  previously fact-checked claims, and fake news.
\newblock In \emph{Proceedings of the European Conference on Information
  Retrieval}, ECIR~'21, pages 639--649, Lucca, Italy. Springer.

\bibitem[{Nakov et~al.(2021{\natexlab{c}})Nakov, Giovanni, Elsayed,
  Barr{\'{o}}n{-}Cede{\~{n}}o, M\'{i}guez, Shaar, Alam, Haouari, Hasanain,
  Mansour, Hamdan, Ali, Babulkov, Nikolov, Shahi, Struß, Mandl, Kutlu, and
  Kartal}]{clef-checkthat:2021:LNCS}
Preslav Nakov, Da~San~Martino Giovanni, Tamer Elsayed, Alberto
  Barr{\'{o}}n{-}Cede{\~{n}}o, Rub\'{e}n M\'{i}guez, Shaden Shaar, Firoj Alam,
  Fatima Haouari, Maram Hasanain, Watheq Mansour, Bayan Hamdan, Zien~Sheikh
  Ali, Nikolay Babulkov, Alex Nikolov, Gautam~Kishore Shahi, Julia~Maria
  Struß, Thomas Mandl, Mucahid Kutlu, and Yavuz~Selim Kartal.
  2021{\natexlab{c}}.
\newblock Overview of the {CLEF}-2021 {CheckThat}! lab on detecting
  check-worthy claims, previously fact-checked claims, and fake news.
\newblock In \emph{Experimental {IR} Meets Multilinguality, Multimodality, and
  Interaction. {P}roceedings of the Twelfth International Conference of the
  {CLEF} Association}, LNCS (12880).

\bibitem[{Nguyen et~al.(2020)Nguyen, Sugiyama, Nakov, and Kan}]{CIKM2020:FANG}
Van-Hoang Nguyen, Kazunari Sugiyama, Preslav Nakov, and Min-Yen Kan. 2020.
\newblock {FANG}: Leveraging social context for fake news detection using graph
  representation.
\newblock In \emph{Proceedings of the 29th ACM International Conference on
  Information and Knowledge Management}, CIKM~'20, pages 1165--1174.

\bibitem[{Nie et~al.(2019)Nie, Chen, and Bansal}]{nie2019combining}
Yixin Nie, Haonan Chen, and Mohit Bansal. 2019.
\newblock Combining fact extraction and verification with neural semantic
  matching networks.
\newblock In \emph{Proceedings of the AAAI Conference on Artificial
  Intelligence}, volume~33 of \emph{AAAI~'19}, pages 6859--6866, Honolulu,
  Hawaii, USA.

\bibitem[{Ostrowski et~al.(2021)Ostrowski, Arora, Atanasova, and
  Augenstein}]{ostrowski2020multi}
Wojciech Ostrowski, Arnav Arora, Pepa Atanasova, and Isabelle Augenstein. 2021.
\newblock Multi-hop fact checking of political claims.
\newblock In \emph{Proceedings of the 25th International Joint Conference on
  Artificial Intelligence}, IJCAI~'21, pages 3892--3898, Motreal, Canada.

\bibitem[{Pritzkau(2021)}]{clef-checkthat:2021:task1:nlytics2021}
Albert Pritzkau. 2021.
\newblock {NLytics} at {CheckThat!} 2021: Check-worthiness estimation as a
  regression problem on transformers.
\newblock In \emph{{CLEF} 2021 Working Notes. {W}orking Notes of {CLEF}
  2021--Conference and Labs of the Evaluation Forum}, Bucharest, Romania
  (online).

\bibitem[{Reimers and Gurevych(2019)}]{reimers-gurevych-2019-sentence}
Nils Reimers and Iryna Gurevych. 2019.
\newblock \href {https://doi.org/10.18653/v1/D19-1410} {Sentence-{BERT}:
  Sentence embeddings using {S}iamese {BERT}-networks}.
\newblock In \emph{Proceedings of the 2019 Conference on Empirical Methods in
  Natural Language Processing and the 9th International Joint Conference on
  Natural Language Processing (EMNLP-IJCNLP)}, pages 3982--3992, Hong Kong,
  China.

\bibitem[{Shaar et~al.(2021{\natexlab{a}})Shaar, Alam, Da~San~Martino, Nikolov,
  Zaghouani, Nakov, and Feldman}]{shaar2021findings}
Shaden Shaar, Firoj Alam, Giovanni Da~San~Martino, Alex Nikolov, Wajdi
  Zaghouani, Preslav Nakov, and Anna Feldman. 2021{\natexlab{a}}.
\newblock \href {https://doi.org/10.18653/v1/2021.nlp4if-1.12} {Findings of the
  {NLP}4{IF}-2021 shared tasks on fighting the {COVID}-19 infodemic and
  censorship detection}.
\newblock In \emph{Proceedings of the Fourth Workshop on NLP for Internet
  Freedom: Censorship, Disinformation, and Propaganda}, pages 82--92, Online.

\bibitem[{Shaar et~al.(2020{\natexlab{a}})Shaar, Babulkov, Da~San~Martino, and
  Nakov}]{shaar-etal-2020-known}
Shaden Shaar, Nikolay Babulkov, Giovanni Da~San~Martino, and Preslav Nakov.
  2020{\natexlab{a}}.
\newblock \href {https://doi.org/10.18653/v1/2020.acl-main.332} {That is a
  known lie: Detecting previously fact-checked claims}.
\newblock In \emph{Proceedings of the 58th Annual Meeting of the Association
  for Computational Linguistics}, pages 3607--3618, Online.

\bibitem[{Shaar et~al.(2021{\natexlab{b}})Shaar, Haouari, Mansour, Hasanain,
  Babulkov, Alam, Da~San~Martino, Elsayed, and
  Nakov}]{clef-checkthat:2021:task2}
Shaden Shaar, Fatima Haouari, Watheq Mansour, Maram Hasanain, Nikolay Babulkov,
  Firoj Alam, Giovanni Da~San~Martino, Tamer Elsayed, and Preslav Nakov.
  2021{\natexlab{b}}.
\newblock Overview of the {CLEF}-2021 {CheckThat}! lab task 2 on detecting
  previously fact-checked claims in tweets and political debates.
\newblock In \emph{Working Notes of CLEF 2021---Conference and Labs of the
  Evaluation Forum}, CLEF~'2021, Bucharest, Romania (online).

\bibitem[{Shaar et~al.(2021{\natexlab{c}})Shaar, Hasanain, Hamdan, Ali,
  Haouari, Nikolov, Kutlu, Kartal, Alam, Da~San~Martino,
  Barr{\'{o}}n{-}Cede{\~{n}}o, M\'{i}guez, Elsayed, and
  Nakov}]{clef-checkthat:2021:task1}
Shaden Shaar, Maram Hasanain, Bayan Hamdan, Zien~Sheikh Ali, Fatima Haouari,
  Alex Nikolov, Mucahid Kutlu, Yavuz~Selim Kartal, Firoj Alam, Giovanni
  Da~San~Martino, Alberto Barr{\'{o}}n{-}Cede{\~{n}}o, Rub\'{e}n M\'{i}guez,
  Tamer Elsayed, and Preslav Nakov. 2021{\natexlab{c}}.
\newblock Overview of the {CLEF}-2021 {CheckThat}! lab task 1 on
  check-worthiness estimation in tweets and political debates.
\newblock In \emph{Working Notes of CLEF 2021---Conference and Labs of the
  Evaluation Forum}, CLEF~'2021, Bucharest, Romania (online).

\bibitem[{Shaar et~al.(2020{\natexlab{b}})Shaar, Nikolov, Babulkov, Alam,
  Barr\'{o}n-Cede{\~n}o, Elsayed, Hasanain, Suwaileh, Haouari, {Da San
  Martino}, and Nakov}]{clef-checkthat-en:2020}
Shaden Shaar, Alex Nikolov, Nikolay Babulkov, Firoj Alam, Alberto
  Barr\'{o}n-Cede{\~n}o, Tamer Elsayed, Maram Hasanain, Reem Suwaileh, Fatima
  Haouari, Giovanni {Da San Martino}, and Preslav Nakov. 2020{\natexlab{b}}.
\newblock Overview of {CheckThat!} 2020 {E}nglish: Automatic identification and
  verification of claims in social media.
\newblock In \emph{Working Notes of CLEF 2020---Conference and Labs of the
  Evaluation Forum}, CLEF~'2020, Thessaloniki, Greece.

\bibitem[{Sheng et~al.(2021)Sheng, Cao, Zhang, Li, and
  Zhong}]{sheng-etal-2021-article}
Qiang Sheng, Juan Cao, Xueyao Zhang, Xirong Li, and Lei Zhong. 2021.
\newblock \href {https://doi.org/10.18653/v1/2021.acl-long.425} {Article
  reranking by memory-enhanced key sentence matching for detecting previously
  fact-checked claims}.
\newblock In \emph{Proceedings of the 59th Annual Meeting of the Association
  for Computational Linguistics and the 11th International Joint Conference on
  Natural Language Processing}, ACL-IJCNLP~'21, pages 5468--5481, Online.

\bibitem[{Shu et~al.(2017)Shu, Sliva, Wang, Tang, and
  Liu}]{Shu:2017:FND:3137597.3137600}
Kai Shu, Amy Sliva, Suhang Wang, Jiliang Tang, and Huan Liu. 2017.
\newblock Fake news detection on social media: A data mining perspective.
\newblock \emph{{ACM SIGKDD} Explorations Newsletter}, 19(1):22--36.

\bibitem[{Stencel(2019)}]{stencel2019number}
Mark Stencel. 2019.
\newblock Number of fact-checking outlets surges to 188 in more than 60
  countries.
\newblock \emph{Duke Reporters’ LAB}, pages 12--17.

\bibitem[{Strubell et~al.(2019)Strubell, Ganesh, and
  McCallum}]{strubell-etal-2019-energy}
Emma Strubell, Ananya Ganesh, and Andrew McCallum. 2019.
\newblock \href {https://doi.org/10.18653/v1/P19-1355} {Energy and policy
  considerations for deep learning in {NLP}}.
\newblock In \emph{Proceedings of the 57th Annual Meeting of the Association
  for Computational Linguistics}, pages 3645--3650, Florence, Italy.

\bibitem[{Thorne and Vlachos(2018)}]{thorne-vlachos:2018:C18-1}
James Thorne and Andreas Vlachos. 2018.
\newblock \href {https://aclanthology.org/C18-1283} {Automated fact checking:
  Task formulations, methods and future directions}.
\newblock In \emph{Proceedings of the 27th International Conference on
  Computational Linguistics}, COLING~'18, pages 3346--3359, Santa Fe, New
  Mexico, USA.

\bibitem[{Thorne et~al.(2018)Thorne, Vlachos, Christodoulopoulos, and
  Mittal}]{thorne-etal-2018-fever}
James Thorne, Andreas Vlachos, Christos Christodoulopoulos, and Arpit Mittal.
  2018.
\newblock \href {https://doi.org/10.18653/v1/N18-1074} {{FEVER}: a large-scale
  dataset for fact extraction and {VER}ification}.
\newblock In \emph{Proceedings of the 2018 Conference of the North {A}merican
  Chapter of the Association for Computational Linguistics: Human Language
  Technologies}, NAACL~'18, pages 809--819, New Orleans, Louisiana.

\bibitem[{Vasileva et~al.(2019)Vasileva, Atanasova, M{\`a}rquez,
  Barr{\'o}n-Cede{\~n}o, and Nakov}]{RANLP2019:checkworthiness:multitask}
Slavena Vasileva, Pepa Atanasova, Llu{\'\i}s M{\`a}rquez, Alberto
  Barr{\'o}n-Cede{\~n}o, and Preslav Nakov. 2019.
\newblock \href {https://doi.org/10.26615/978-954-452-056-4_141} {It takes nine
  to smell a rat: Neural multi-task learning for check-worthiness prediction}.
\newblock In \emph{Proceedings of the International Conference on Recent
  Advances in Natural Language Processing}, RANLP~'19, pages 1229--1239, Varna,
  Bulgaria.

\bibitem[{Vlachos and Riedel(2014)}]{vlachos2014fact}
Andreas Vlachos and Sebastian Riedel. 2014.
\newblock Fact checking: Task definition and dataset construction.
\newblock In \emph{Proceedings of the ACL 2014 Workshop on Language
  Technologies and Computational Social Science}, pages 18--22, Baltimore,
  Maryland, USA.

\bibitem[{Vo and Lee(2018)}]{vo2018rise}
Nguyen Vo and Kyumin Lee. 2018.
\newblock \href {https://doi.org/10.1145/3209978.3210037} {The rise of
  guardians: Fact-checking {URL} recommendation to combat fake news}.
\newblock In \emph{Proceedings of the 41st International ACM SIGIR Conference
  on Research \& Development in Information Retrieval}, SIGIR '18, pages
  275--284, Ann Arbor, Michigan, USA.

\bibitem[{Vo and Lee(2020)}]{vo-lee-2020-where-facts}
Nguyen Vo and Kyumin Lee. 2020.
\newblock \href {https://doi.org/10.18653/v1/2020.emnlp-main.621} {Where are
  the facts? {S}earching for fact-checked information to alleviate the spread
  of fake news}.
\newblock In \emph{Proceedings of the 2020 Conference on Empirical Methods in
  Natural Language Processing}, EMNLP~'20, pages 7717--7731, Online.

\bibitem[{Vosoughi et~al.(2018{\natexlab{a}})Vosoughi, Roy, and
  Aral}]{Vosoughi1146}
Soroush Vosoughi, Deb Roy, and Sinan Aral. 2018{\natexlab{a}}.
\newblock The spread of true and false news online.
\newblock \emph{Science}, 359(6380):1146--1151.

\bibitem[{Vosoughi et~al.(2018{\natexlab{b}})Vosoughi, Roy, and
  Aral}]{vosoughi2018spread}
Soroush Vosoughi, Deb Roy, and Sinan Aral. 2018{\natexlab{b}}.
\newblock The spread of true and false news online.
\newblock \emph{Science}, 359(6380):1146--1151.

\bibitem[{Wadden et~al.(2020)Wadden, Lin, Lo, Wang, van Zuylen, Cohan, and
  Hajishirzi}]{wadden-etal-2020-fact}
David Wadden, Shanchuan Lin, Kyle Lo, Lucy~Lu Wang, Madeleine van Zuylen, Arman
  Cohan, and Hannaneh Hajishirzi. 2020.
\newblock \href {https://doi.org/10.18653/v1/2020.emnlp-main.609} {Fact or
  fiction: Verifying scientific claims}.
\newblock In \emph{Proceedings of the 2020 Conference on Empirical Methods in
  Natural Language Processing (EMNLP)}, pages 7534--7550, Online.

\bibitem[{Wang(2017)}]{wang:2017:Short}
William~Yang Wang. 2017.
\newblock \href {https://doi.org/10.18653/v1/P17-2067} {{``L}iar, liar pants on
  fire{''}: A new benchmark dataset for fake news detection}.
\newblock In \emph{Proceedings of the 55th Annual Meeting of the Association
  for Computational Linguistics}, ACL~'17, pages 422--426, Vancouver, Canada.

\bibitem[{Yoneda et~al.(2018)Yoneda, Mitchell, Welbl, Stenetorp, and
  Riedel}]{yoneda2018ucl}
Takuma Yoneda, Jeff Mitchell, Johannes Welbl, Pontus Stenetorp, and Sebastian
  Riedel. 2018.
\newblock \href {https://doi.org/10.18653/v1/W18-5515} {{UCL} machine reading
  group: Four factor framework for fact finding ({H}exa{F})}.
\newblock In \emph{Proceedings of the First Workshop on Fact Extraction and
  {VER}ification}, FEVER~'18, pages 97--102, Brussels, Belgium.

\bibitem[{Zaman et~al.(2014)Zaman, Fox, Bradlow et~al.}]{zaman2014bayesian}
Tauhid Zaman, Emily~B Fox, Eric~T Bradlow, et~al. 2014.
\newblock A bayesian approach for predicting the popularity of tweets.
\newblock \emph{Annals of Applied Statistics}, 8(3):1583--1611.

\bibitem[{Zhao et~al.(2019)Zhao, Xiong, Rosset, Song, Bennett, and
  Tiwary}]{zhao2019transformer}
Chen Zhao, Chenyan Xiong, Corby Rosset, Xia Song, Paul Bennett, and Saurabh
  Tiwary. 2019.
\newblock {Transformer-XH}: Multi-evidence reasoning with extra hop attention.
\newblock In \emph{Proceedings of the International Conference on Learning
  Representations}, ICLR~'19, New Orleans, Louisiana, USA.

\bibitem[{Zhong et~al.(2020)Zhong, Xu, Tang, Xu, Duan, Zhou, Wang, and
  Yin}]{zhong2020reasoning}
Wanjun Zhong, Jingjing Xu, Duyu Tang, Zenan Xu, Nan Duan, Ming Zhou, Jiahai
  Wang, and Jian Yin. 2020.
\newblock \href {https://doi.org/10.18653/v1/2020.acl-main.549} {Reasoning over
  semantic-level graph for fact checking}.
\newblock In \emph{Proceedings of the 58th Annual Meeting of the Association
  for Computational Linguistics}, ACL~'20, pages 6170--6180, Online.

\bibitem[{Zhou et~al.(2019)Zhou, Han, Yang, Liu, Wang, Li, and
  Sun}]{zhou2019gear}
Jie Zhou, Xu~Han, Cheng Yang, Zhiyuan Liu, Lifeng Wang, Changcheng Li, and
  Maosong Sun. 2019.
\newblock \href {https://doi.org/10.18653/v1/P19-1085} {{GEAR}: Graph-based
  evidence aggregating and reasoning for fact verification}.
\newblock In \emph{Proceedings of the 57th Annual Meeting of the Association
  for Computational Linguistics}, ACL~'19, pages 892--901, Florence, Italy.

\end{thebibliography}

% \newpage
% \clearpage
% \section*{Appendix}
% \label{sec:appendix}
% \appendix
% \input{sections/supplemental_material}

\end{document}